\documentclass{article}

\usepackage{arxiv}

\usepackage[utf8]{inputenc} 
\usepackage[T1]{fontenc}    
\usepackage{hyperref}       
\usepackage{url}            
\usepackage{booktabs}       
\usepackage{amsfonts}       
\usepackage{nicefrac}       
\usepackage{microtype}      
\usepackage{lipsum}
\usepackage{graphicx}
\usepackage{colortbl}
\usepackage{biblatex}
\usepackage{subcaption}
\usepackage{float}
\usepackage{cleveref}
\bibliography{references.bib}
\graphicspath{ {./images/} }

\title{Adaptive Explicit Kernel Minkowski Weighted K-means}

\author{
 Amir Aradnia \\
  Department of Computer Engineering\\
  Amirkabir University of Technology\\
   Tehran, Iran \\
  \texttt{a.mahdi@aut.ac.ir} \\
   \And
Maryam Amir Haeri \\
 Learning, Data-Analytics and Technology Department\\
  University of Twente \\
  Enschede, Netherlands\\
  \texttt{m.amirhaeri@utwente.nl} \\
  \And
Mohammad Mehdi Ebadzadeh \\
  Department of Computer Engineering\\
  Amirkabir University of Technology\\
   Tehran, Iran \\
  \texttt{ebadzadeh@aut.ac.ir} \\
 }
\usepackage[english]{babel}
 
\usepackage{biblatex}
\begin{document}
\maketitle
\begin{abstract}
The K-means algorithm is among the most commonly used data clustering methods. However, the  regular K-means can only be applied in the input space and it is applicable when clusters are linearly separable. The kernel K-means, which extends K-means into the kernel space, is able to capture nonlinear structures and identify arbitrarily shaped clusters. However, kernel methods often operate on the kernel matrix of the data, which scale poorly with the size of the matrix or suffer from the high clustering cost due to the  repetitive calculations of kernel values. Another issue is that algorithms access the data only through evaluations of $K(x_i, x_j)$, which limits many processes that can be done on data through the clustering task. This paper proposes a method to combine the advantages of the linear and nonlinear approaches by using driven corresponding approximate finite-dimensional feature maps based on spectral analysis. Applying approximate finite-dimensional feature maps were only discussed in the Support Vector Machines (SVM) problems before. We suggest using this method in kernel K-means era as alleviates storing huge kernel matrix in memory, further calculating cluster centers more efficiently and access the data explicitly in feature space. These explicit feature maps enable us to access the data in the feature space explicitly and take advantage of K-means extensions in that space. We demonstrate our Explicit Kernel Minkowski Weighted K-mean (Explicit KMWK-mean) method is able to  be more adopted and find best-fitting values in new space by applying additional Minkowski exponent and feature weights parameter.  Moreover, it can reduce the impact of  concentration on nearest neighbour search by suggesting  investigate among other norms instead of Euclidean norm, includes Minkowski norms and fractional norms (as an extension  of the Minkowski norms with p<1). The proposed method is evaluated by four benchmark data sets and compared the performance with commonly used kernel clustering approach. Experiments show the proposed method consistently achieves superiors clustering performances while it can be used to reduce memory consumption as well.
\end{abstract}

\keywords{Kernel clustering \and Minkowski metric \and Features map \and K-means}

\section{Introduction}
Clustering can be considered as the most important unsupervised learning problem. Clustering methods are used to determine the intrinsic grouping in a set of unlabeled data. It has been shown to be useful in many practical domains such as web search, image segmentation,  image compression, gene expression analysis, recommendation systems and mining text data  \cite{daruru2009pervasive,jiang2004cluster,dubes1980clustering}. K-means clustering \cite{macqueen1967some} is one of the most popular conventional clustering algorithms, despite its age. It aims to partition sample of \textit{N} observation into \textit{K} compact clusters in an iterative process. The K-means algorithm only works reasonably well when  1) clusters can be separated by hyper-planes and 2) each data point belongs to the closest cluster center. If one of these principles does not hold, the standard K-means algorithm will likely not give a good result. Kernel-based clustering methods overcome these limitations by using an appropriate non-linear mapping to higher dimensional feature space. Thus, it enables the K-means algorithm to partition data points by the linear separator in the new space, that has non-linear projection back in the original space. Various types of kernel-based methods such as the kernel version of the SOM (Self-organizing map) algorithm \cite{macdonald2000kernel,inokuchi2004lvq}, kernel neural gas\cite{qin2004kernel}, one Class SVM (Support Vector Machines) \cite{camastra2005novel} and kernel fuzzy clustering \cite{zhang2004novel,zhang2003kernel} have been proposed by researchers. In this paper, we focus on kernel k-means clustering because of its efficiency and simpleness. Furthermore, various studies \cite{zha2002spectral,ding2005equivalence,dhillon2004unified} claim that different kernel-based clustering methods show similar result as kernel K-means.
\par
Although kernel-based methods have received considerable attention from the machine learning community in recent years, they still suffer from the following problems in real applications. First, the high clustering cost due to either the repeated calculations of kernel values or requiring huge amounts of memory to store the kernel matrix makes it unsuitable for large corpora. Second, algorithms access the data only through evaluations of \textit{K(x, y)}. Therefore, many processes on the data points like fighting the concentration phenomenon, handling noise, etc. limited to work in the original space.
\par
 
The aim of this paper is to develop a clustering method that can group data points with both linear and non-linear structures while try to address the two mentioned problems of kernel clustering methods. As the main contribution of this paper, we address both the space complexity of storing kernel matrix and lack of accessing to data in feature space by proposing  new \emph{Adaptive Explicit Kernel Minkowski Weighted K-means (Explicit KMWK-mean)} method. The proposed method combines the advantages of the linear and nonlinear approaches by using driven corresponding approximate finite-dimensional feature maps based on 1D Fourier analysis \cite{vedaldi2012efficient} for a large family of additive kernels, known as $\gamma$-homogeneous kernels. The proposed method, first map data to feature space. This feature space is data independent approximation of  $\gamma$-homogeneous kernels. Then, in order to provide a better fit to cluster structure, the weighted version of  Minkowski K-means in low-dimensional feature space would be applied. Especially, we analyze the concentration of the Euclidean norm and the impact of distance measure on concentration and investigate on Minkowski norms and fractional norms, an extension of the Minkowski norms with $p <1$, as a measure of distance among data in the kernel space. Adaptive Minkowski metric allows to fight against possible concentration in high-dimensional spaces and the weighting property enables our algorithm to cover spherical and non-spherical (elliptical) structures in feature space and getting the chance of finding even more complex clusters in feature space.

\par
The remainder of this article is organized as follows. Section 2, briefly describes k-means and kernel k-means as the preliminary notions. Section 3 introduces Explicit Feature Maps and especially focused on Homogeneous kernels. Section 4 presents our modified version of kernel K-means by analyzing the alternative Minkowski distance for arbitrary $p\in \mathbb{R}^+$ instead of Euclidean one in feature space. In Section 5 we describe the results of experiments four benchmark data sets.  Section 6 concludes the paper.

\section{K-means and Kernel K-means}
The K-means method is designed to partition $N$ D-dimensional samples $X=(x_{1},x_{2},\ldots,x_{N}).$ in to $K$ clusters ${C_1}{\rm{, }}{C_2}{\rm{, }}...{\rm{, }}{C_K}$ and return centroid vector for each cluster $M=(m_{1},m_{2},\ldots,m_{K}).$ The batch mode K-means algorithm would operate by the following iterative procedure:
\begin{enumerate}
  \item Initialize $K$ cluster center $m_{1},m_{2},\ldots,m_{K}.$
  \item Assign each sample $x_{i}$ to its closest center. Namely, compute the indicator matrix ${\delta _{ik}},\:(1 \le k \le K).$ 
  \begin{equation}\label{indicator_eqn}
{\delta _{ik}}{\rm{ = }}\left\{ {\begin{array}{*{20}{c}}
1&{d({{\rm{x}}_{\rm{i}}},{{\rm{m}}_{\rm{k}}}) < d({{\rm{x}}_{\rm{i}}},{{\rm{m}}_{\rm{j}}})\:{\rm{for}}\:{\rm{all}}\:{\rm{j}} \ne k}\\
0&{otherwise}
\end{array}} \right.
\end{equation}
 \item Update the cluster centers.
  \begin{equation}\label{Euc_center}
{m_k} = \frac{1}{{|{c_k}|}}\sum\limits_{i = 1}^N {{\delta _{ik}}} {x_i}
  \end{equation}
  where $|C_{k}|$ is the number of samples in $C_{k}$.
  \item Iterate between (2) and (3) until convergence.
  \item Return $m_{1},m_{2},\ldots,m_{K}.$
\end{enumerate}
Note that, in (\ref{indicator_eqn}), $d(x_{i},m_{k})$ is the Euclidean distance given by:
\begin{equation}
{d^2}({x_i},{m_k}) = ||{x_i} - {m_k}|{|^2}
\end{equation}
The preceding procedure actually is an iterative solution to optimization problem that attempts to minimize the objective function as follows:
\begin{equation} \label{obj_func}
\arg {\rm{ }}\mathop {{\rm{min}}}\limits_{M,\delta } \sum\limits_{i = 1}^N {\sum\limits_{k = 1}^K {{\delta _{ik}}||{x_i} - {m_k}|{|^2}} } 
\end{equation}

In most cases, the distance in use is the squared Euclidean distance. However, the Euclidean distances tend to concentrate, when data are high dimensional. This means that, all the pairwise distances may converge to the same value. Accordingly, the relevance of Euclidean distance has been questioned in the past and alternative norms, especially fractional norms ($L_{p}$  semi-norms with $p<1$) were suggested to reduce concentration phenomenon \cite{hinneburg2000nearest,aggarwal2001surprising}. Obviously, by using different metrics,  cluster centers don’t follow from the equation in (\ref{Euc_center}) anymore. Finding fractional and Minkowski’s centers, whose components are minimizers of the summary corresponding distances, discussed in Section \ref{finding_centers}.
\par
The K-means algorithm with Euclidean distance generally works on ellipse-shaped clusters. It is not applicable when elliptical regions does not hold. By applying some kind of transformation to the data, mapping them to some new space, the k-means algorithm may be achieve better   performance than in original space. Again, suppose we are given N samples of $X=(x_{1},x_{2},\ldots,x_{N})\:x_{i}\in \mathbb{R}^{D}$, and mapping function $\Phi$ that transforms $x_{i}$ from original space $\mathbb{R}^{D}$ to a high dimensional feature space $\mathbb{H}$. Kernel functions are implicitly defined as the dot product of two vectors in the new transformed feature space.
\begin{equation}
K(x_{i},x_{j}) = \Phi (x_{i}).\Phi (x_{j})  
\end{equation}
In the rest of paper, we use ${\Phi _i}$ instead of $\Phi ({x_i})$ for ease of description. Essentially, the  transformation is defined implicitly, without knowing the concrete form of $\Phi$ \cite{shawe2000support}. Computation of distances in the transformed space is one of the most important issues when K-means is extended to the kernel K-means. The squared Euclidean distance between $x_{i}$ and $x_{j}$ in feature space would be as:
\begin{equation}
\begin{array}{l}
\begin{array}{l}
\mathop {{d^2}}\limits_{Euc}({\Phi _i},{\Phi _j}) = ||{\Phi _i} - {\Phi _j}|{|^2} = {\Phi _i}^2 - 2{\Phi _i}.{\Phi _j} + {\Phi _j}^2\\
 = K({x_i},{x_i}) - 2K({x_i},{x_j}) + K({x_j},{x_j})
\end{array}
\end{array}
\end{equation}
The cluster center in transformed space can be calculated as given below:

\begin{equation}
{m'_k} = \frac{1}{{|{C_k}|}}\sum\limits_{i = 1}^N {{\delta _{ik}}} {\Phi _i}
\end{equation}
Therefore, the distance of each data point and the cluster center in new space can be computed without knowing the transformation $\Phi$ explicitly.
\begin{equation}\label{kerneldist_eqn}
\begin{array}{*{20}{l}}
{d{^2}({\Phi _i},{m'_k}) = ||{\Phi _i} - \frac{1}{{|{C_k}|}}\sum\limits_{j = 1}^N {{\delta '_{jk}}.} {\Phi _j}|{|^2}}\\
{ = K({x_i},{x_i}) - \frac{2}{{|{C_k}|}}\sum\nolimits_{{x_{j \in {C_k}}}} {K({x_i},{x_j})}  + \frac{1}{{|{C_k}{|^2}}}\sum\nolimits_{{x_{j \in {C_k}}}} {\sum\nolimits_{{x_{l \in {C_k}}}} {K({x_i},{x_l})} } }
\end{array}
\end{equation}
Where ${\delta'}$ is the indicator matrix which ${{\delta '_{jk}}}$ indicates whether ${{\Phi _i}}$ is assigned to ${{C_k}}$
or not. We will be moved from K-means to Kernel K-means by applying(\ref{kerneldist_eqn}) to the standard K-means. \par
The following notation is  used in the rest of the paper. Multidimensional additive kernels  have been represented as $K(x_i,x_j)$, henceforth $k(x_i,x_j)$ will be used  for the scalar ones. Multidimensional kernel function can be obtained from scalar ones as: $K({x_i},{x_j}) = \sum\limits_{l = 1}^D k (x_i^l,x_j^l)$. The scalar feature map is also denoted by $\varphi ({x_i})$ and multidimensional ones are given by ${\Phi _i} = \mathop  \oplus \limits_{l = 1}^D \varphi _i^l$, it means ${\Phi _i} = [\varphi _i^1,...\varphi _i^D]$.

\section{Explicit Feature Maps }
Namely, in kernel learning context, for each positive-definite (PD) kernel $K(x_i,x_j)$ there exists a corresponding mapping function $\Phi$ to an arbitrary dimensional space such that $K(x_i,x_j)=\Phi_i.\Phi_j.$ Even though the explicit form of the mapping function is useful conceptually, it is not often used in computations. Typically, these feature spaces are infinite-dimensional, yet it is possible to find an efficient finite-dimensional approximation of $\Phi$. In the following, we will introduce a class of kernels which commonly used in computer vision, called homogeneous kernels. Then describe deriving corresponding approximate explicit finite-dimensional feature maps proposed by Vedaldi and Zisserman \cite{vedaldi2012efficient}. The main focus of this paper, as in an influential paper \cite{vedaldi2012efficient}, is a class of additive kernels such as the Hellinger’s, ${\chi ^2}$, intersection, and Jensen-Shannon ones, which are frequently used in computer vision applications. All of the mentioned kernels are mutual in two properties of additivity and homogeneity. Common homogeneous kernels with their properties are listed in table \ref{tab:kernels_feature}.

\begin{table}
\centering
\caption{Well-known kernel functions and their corresponding distance , signature and closed form feature maps. This table is adopted from \cite{vedaldi2012efficient}.}
\begin{tabular}{l|lllll}
Kernel&    $K(x,y)$    & $d(x,y)$ & Signature$\kappa (\lambda )$   & $\rho (\omega )$   & ${\varphi _\omega }({x^i})$   \\ 
\hline
$\chi ^2$ &    $\sum {\frac{{{{({x^i} - {y^i})}^2}}}{{({x^i} - {y^i})}}} $            & $\sum {\frac{{2{x^i}{y^j}}}{{({x^i} + {y^j})}}} $      & ${\rm{ sech(}}\lambda /2{\rm{)}}$    & ${\mathop{\rm sech}\nolimits} (\pi \omega )$    & ${e^{i\omega {\rm{ log }}{{\rm{x}}^i}}}\sqrt {{x^i}{\rm{ sech(}}\pi \omega {\rm{)}}}$    \\ 
\hline
Hellinger         & $2\sum {(\sqrt {{x^i}} }  - \sqrt {{y^i}} {)^2}$         & $\sum {\sqrt {{x^i}{y^i}} } $     & $1$   & $\delta (\omega )$   & $\sqrt {{x^i}} $   \\ 
\hline
JS        & $\begin{array}{c}
\sum {({x^i}{{\log }_2}(\frac{{2{x^i}}}{{{x^i} + {y^j}}})}  + \\
{y^i}{\log _2}(\frac{{2{y^i}}}{{{x^i} + {y^j}}}))
\end{array}$          & $\begin{array}{c}
\frac{1}{2}\sum {({x^i}{{\log }_2}(\frac{{{x^i} + {y^j}}}{{{x^i}}})}  + \\
{y^j}{\log _2}(\frac{{{x^i} + {y^j}}}{{{y^j}}}))
\end{array}$    & $\begin{array}{c}
\frac{{{e^{\frac{\lambda }{2}}}}}{2}{\log _2}(1 + {e^{ - \lambda }}) + \\
\frac{{{e^{\frac{{ - \lambda }}{2}}}}}{2}{\log _2}(1 + {e^\lambda })
\end{array}$  & $\frac{2}{{\log 4}}\frac{{{\mathop{\rm sech}\nolimits} (\pi \omega )}}{{1 + 4{\omega ^2}}}$  & ${e^{i\omega {\rm{ log }}{{\rm{x}}^i}}}\sqrt {\frac{2}{{\log 4}}\frac{{{\mathop{\rm sech}\nolimits} (\pi \omega )}}{{1 + 4{\omega ^2}}}} $   \\ 
\hline
intersection       & 
$\sum {|{x^i} - {y^i}|} $         & $\sum {\min ({x^i},{y^i})} $   & ${e^{\frac{{ - |\lambda |}}{2}}}$ & $\frac{2}{{\log 4}}\frac{{{\mathop{\rm sech}\nolimits} (\pi \omega )}}{{1 + 4{\omega ^2}}}$ & ${e^{i\omega {\rm{ log }}{{\rm{x}}^i}}}\sqrt {\frac{2}{\pi }\frac{1}{{1 + 4{\omega ^2}}}} $
\end{tabular}
\label{tab:kernels_feature}
\end{table}

\paragraph{Homogeneous Kernels.} A kernel $k_h(a,b)$ called a homogeneous of degree $\gamma$ if
\begin{equation}
    \forall t \ge 0:{K_h}(ta,tb) = {t^\gamma }{K_h}(a,b)
\end{equation}
Here $a,b \in \mathbb{R}$. If we set $t = \frac{1}{{\sqrt {ab} }}$, the homogeneous kernel can be rewritten as :
\begin{equation}
    \begin{array}{l}
{k_h}(a,b) = {t^{ - \gamma }}{k_h}(ta,tb) = {(ab)^{\frac{\gamma }{2}}}{k_h}(\sqrt {\frac{b}{a}} ,\sqrt {\frac{a}{b}} )\\
 = {(ab)^{\frac{\gamma }{2}}}\kappa (\log b - \log a)
\end{array}
\end{equation}
The scalar function $\kappa $ is called signature function  which is used to deriving associated feature map of homogeneous kernel functions. It is defined as :
\begin{equation}
\kappa (\lambda ) = {k_h}({e^{\frac{\lambda }{2}}},{e^{ - \frac{\lambda }{2}}}),{\rm{ }}\lambda  \in {\rm{R,}}
    \end{equation}
Bochner’s theorem by using Fourier transform of signature function, $\kappa (\lambda )$ can address the problem of which mapping function made the given homogeneous kernel. As a result, the feature map $\varphi $ for $\gamma$-homogeneous kernels will be derived as:
\begin{equation}\label{ex_feature_maps}
    {\varphi _\omega }(a) = {e^{ - i\omega \log a}}\sqrt {{a^\gamma }\rho (\omega )} 
\end{equation}
where $\omega $ can be viewed as the index of vector  dimension and   ${\rho (\omega )}$ is the spectrum function and can be obtained from inverse Fourier transform of the signature $\kappa (\lambda )$.
\begin{equation}
 \rho (\omega ) = \frac{1}{{{{(2\pi )}^D}}}\int_{{R^D}} {{e^{ - i(\omega ,\lambda )}}} \kappa (\lambda )d\lambda .   
\end{equation}

In (\ref{ex_feature_maps}) feature maps are continuous functions but still finite approximation feature maps can be generated by sampling the continuous spectrum and rescaling it. Closed-form of common kernel feature maps are described in table \ref{tab:kernels_feature}. Moreover, Hein and Bousquet \cite{hein2004hilbertian} introduced a large class of $\gamma$-homogeneous Hilbertian metrics and correspondent kernels which encompasses all previously described kernels in table \ref{tab:kernels_feature}. This class of metrics defined as follows with two parameters $\alpha$ and $\beta$ to be tuned.

\begin{equation} \label{general_homoge}
    d_{\alpha |\beta }^2(a,b) = \frac{{{2^{\frac{1}{\beta }}}{{({a^\alpha } + {b^\alpha })}^{\frac{1}{\alpha }}} - {2^{\frac{1}{\alpha }}}{{({a^\beta } + {b^\beta })}^{\frac{1}{\beta }}}}}{{{2^{\frac{1}{\alpha }}} - {2^{\frac{1}{\beta }}}}}
\end{equation}
Function $d$ in Equation (\ref{general_homoge}) is a $\gamma$-homogeneous metric on ${\mathbb{R}^ + }$ with $\alpha  \in [1,\infty ]$ and $\beta  \in [\frac{1}{2},\alpha ]$ or $\beta  \in [ - \infty , - 1]$. The pointwise limit of $d$ when $\alpha  \to \beta $ is determined as:
\begin{equation}
    \begin{array}{l}
\lim {\rm{ d}}_{\alpha |\beta }^2(a,b) = \frac{{{\beta ^2}{2^{\frac{1}{\beta }}}}}{{\log (2)}}\frac{\partial }{{\partial \beta }}{(\frac{{{a^\beta } + {b^\beta }}}{2})^{\frac{1}{\beta }}}\\
 = {(\frac{{{a^\beta } + {b^\beta }}}{{\log (2)}})^{\frac{1}{\beta }}}\left[ {\frac{{{a^\beta }}}{{{a^\beta } + {b^\beta }}}\log (\frac{{2{a^\beta }}}{{{a^\beta } + {b^\beta }}}) + \frac{{{b^\beta }}}{{{b^\beta } + {b^\beta }}}\log (\frac{{2{b^\beta }}}{{{a^\beta } + {b^\beta }}})} \right]
\end{array}
\end{equation}
Corresponding PD kernel for the above can be taken by:
\begin{equation}
k(a,b) = \frac{1}{2}( - {d^2}(a,b) + {d^2}(a,0) + {d^2}(b,0))
\end{equation}
\section{Adaptive Explicit Kernel K-means}
In this section, we present the \emph{Adaptive Explicit Kernel K-means} method for homogeneous kernels. Homogeneous kernels have been introduced in the previous section and a data-independent method \cite{vedaldi2012efficient} for deriving approximate finite-dimensional feature maps was described. This class of kernels is a frequently-used measure for histogram image comparison due to its effectiveness rather than linear kernel(Euclidean distance). Using explicit feature maps alleviates issues around the  storing   huge   similarity   matrix   or   repeated calculation  of kernel  values. Moreover, we use the explicit feature map to take advantage of accessing through the data in feature space. We expect good matching between K-means model and real data structure of data in transformed space by choosing suitable kernel, but we still have this chance to get better fit by extending from the squared Euclidean to arbitrary weighted Minkowski metric in new space. In addition to adding more flexibility, adaptive Minkowski metric allows to fight against possible concentration in high-dimensional spaces. Furthermore, weighting property  with reflect to within-cluster feature variances,enables our algorithm to cover spherical and non-spherical (elliptical) structures. In this way, features with smaller within-cluster variances receive a larger weight and features which more evenly distributed across the cluster get a smaller weight. 
\subsection{Minkowski Metric}
The distance we consider here is the Minkowski distance. It can be seen as a generalization of the Euclidean distance, which is defined as below:
\begin{equation}
   {d_{Mink}}({x_i},{x_j}) = (\sum\limits_{l = 1}^D {|x_i^l - x_j^l{|^p}{)^{\frac{1}{p}}}} 
\end{equation}
Note that, when $0<p<1$  does not hold metrics properties therefore it is not actual metric and the corresponding norm are not  are indeed norms. They are not satisfying the triangle inequality. They are usually called prenorms or fractional norms.

\subsubsection{Concentration in high-dimensional spaces}
Losing discriminative power of Euclidean distance to indexing points in high-dimensional spaces had been shown in the past. This means, as dimension increases, the distance to the nearest point appear to be as the same as the farthest one. This phenomena is known as concentration of distances. This inability of Euclidean distance to distinguish distances in high dimensions caused to  alternative distance measures were suggested. In \cite{hinneburg2000nearest} a theoretical analysis of  absolute difference between the farthest point distance $d_{max}$ and the closest point distance $d_{min}$ for Minkowski norms is presented. It point out that for D-dimensional i.i.d random vectors $x_i , (1 \le i \le N) $
\begin{equation}\label{hinneburg}
C \le \mathop {\lim }\limits_{d \to \infty } {\rm{E}}(\frac{{{{\max }_i}||{x_i}|{|_p} - {{\min }_i}||{x_i}|{|_p}}}{{{D^{\frac{1}{p} - \frac{1}{2}}}}}) \le (N - 1).C 
\end{equation}
Where C is some constant independent of the distribution
of the $x_i$. This means the contrast between closest and farthest neighbor on average grows as $D^{\frac{1}{p} - \frac{1}{2}}$. The authors concluded that $l_1$ and $l_2$ norm may be more relevant than $l_p$ when $p \ge 3$. In fact for $l_p$ with $(p \ge 3)$, the difference between to the farthest and nearest neighbor goes to $0$ as dimensionality increases. These results encouraged researchers to examine fractional distances, $l_p$ distance with $p \in (0,1)$. In \cite{aggarwal2001surprising} authors extended previous work and proposed using fractional distance metric. They has been shown that fractional distance can provide higher relative contrast and meaningful result under same conditions as in (\ref{hinneburg}).\\
The obtained results in \cite{hinneburg2000nearest} and \cite{aggarwal2001surprising} cannot be used in general when the data are not uniformly distributed. Indeed, it is quite rare that data spread through such spaces.observation of real data  shows that high dimensional spaces are mostly empty. In these spaces, it is  common that data spread on a sub-manifold. In \cite{francois2007concentration} data distribution instead of data set has been studied. For that purpose the relative variance ratio is proposed which is defined as follows:
\begin{equation}
    R{V_{F,p}} = \frac{{\sqrt {Var(||x_i|{|_p})} }}{{E(||x_i|{|_p})}}
\end{equation}

 Similarly to the relative contrast (\ref{hinneburg}), the relative variance can be seen as a measure of concentration. Smaller values of $ R{V_{F,p}}$ indicate less concentration. We can  see that the shape of distribution $F$ and the value of $p$ might affect the value of the $R{V_{F,p}}$. Therefore, for adjusting the value of $p$, the shape of distribution should be considered too. It is completely possible that the higher relative variance acquired by higher-order norms.
\subsubsection{Choosing the Optimal Value of p}
\begin{figure}[htb]
    \centering 
\begin{subfigure}{0.30\textwidth}
  \includegraphics[width=\linewidth]{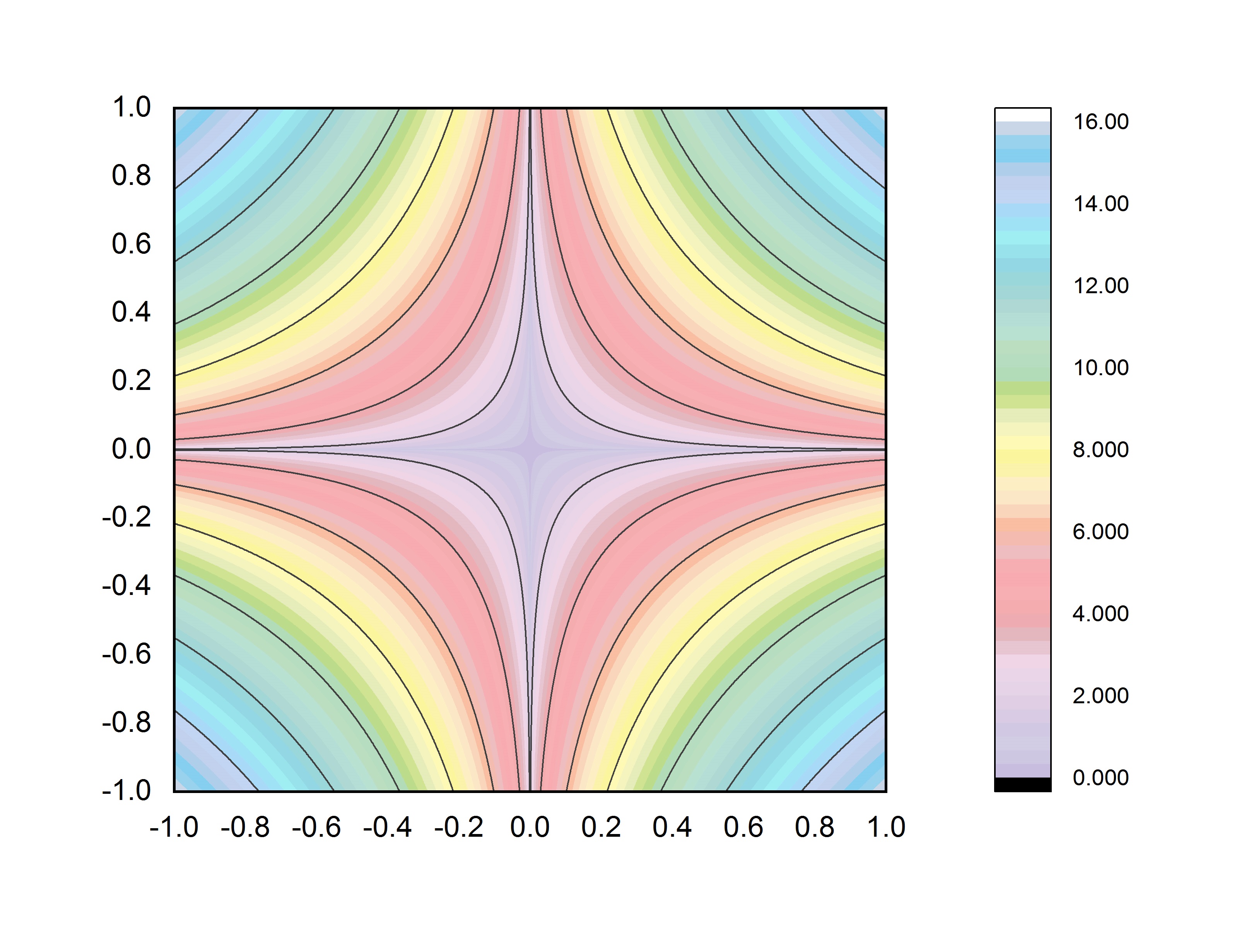}
  \caption{p=0.25}
  \label{fig:p_25}
\end{subfigure}\hfil 
\begin{subfigure}{0.30\textwidth}
  \includegraphics[width=\linewidth]{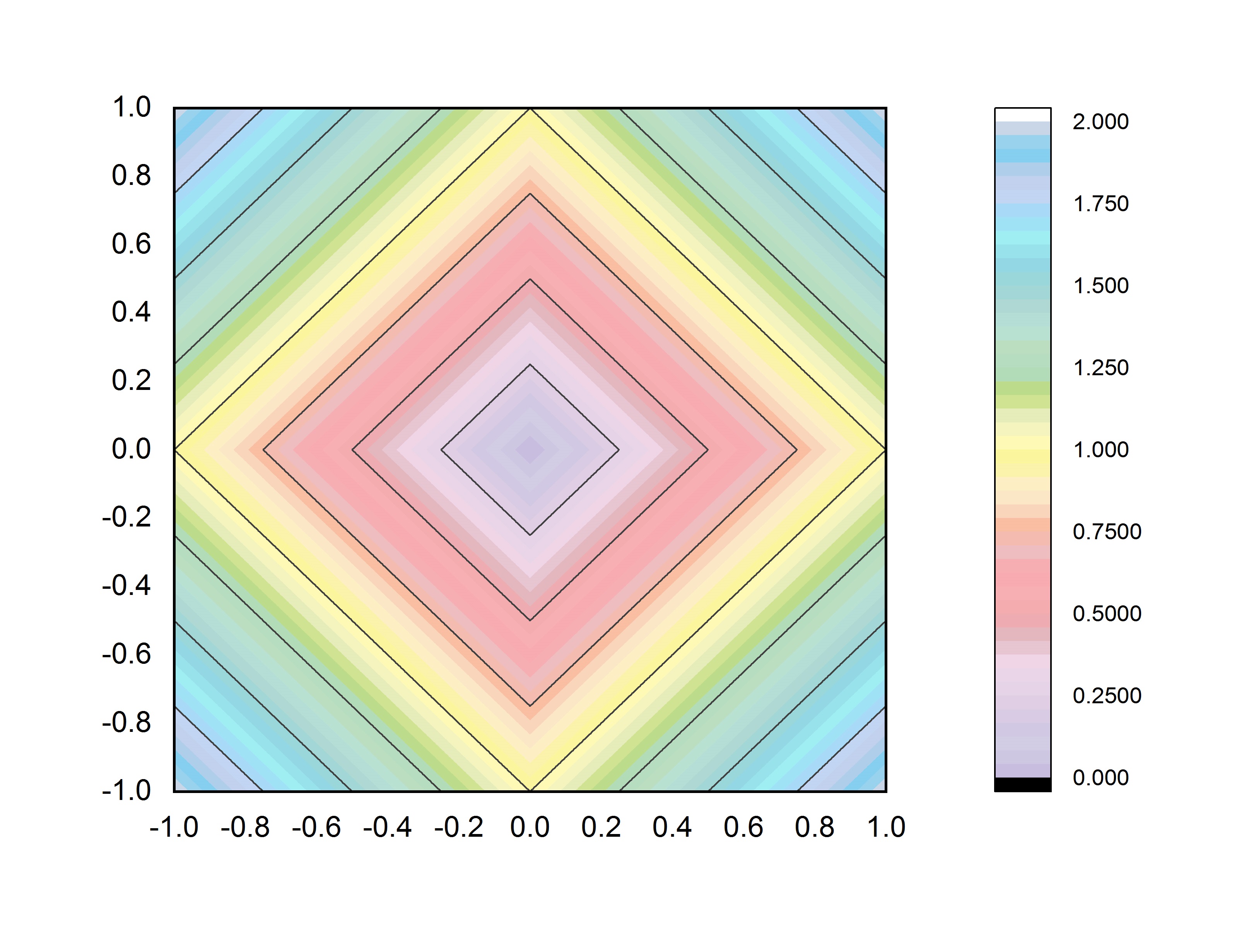}
  \caption{p=1}
  \label{fig:p1}
\end{subfigure}\hfil 
\begin{subfigure}{0.30\textwidth}
  \includegraphics[width=\linewidth]{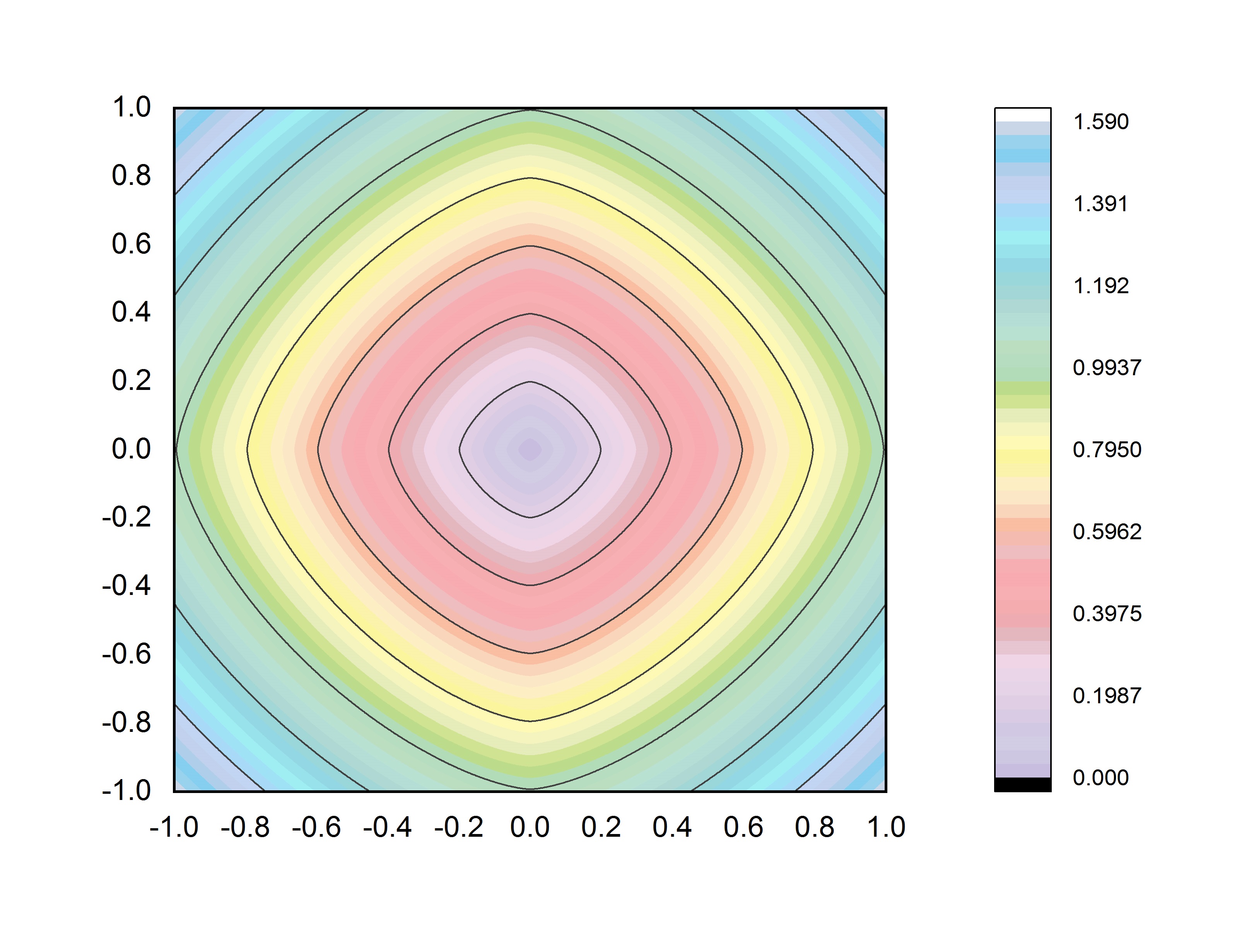}
  \caption{p=1.5}
  \label{fig:p1_5}
\end{subfigure}
\medskip
\begin{subfigure}{0.30\textwidth}
  \includegraphics[width=\linewidth]{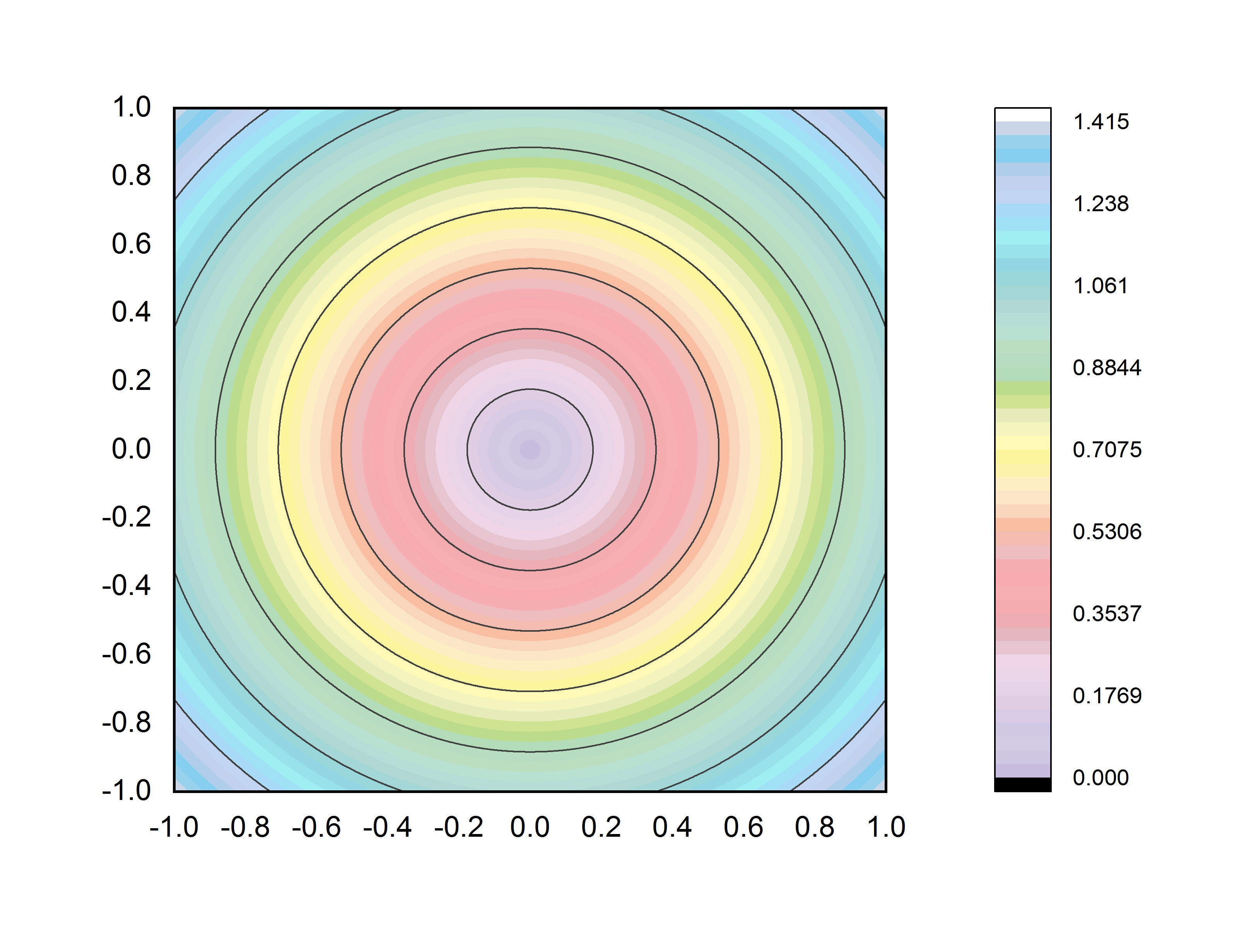}
  \caption{p=2}
  \label{fig:p2}
\end{subfigure}\hfil 
\begin{subfigure}{0.30\textwidth}
  \includegraphics[width=\linewidth]{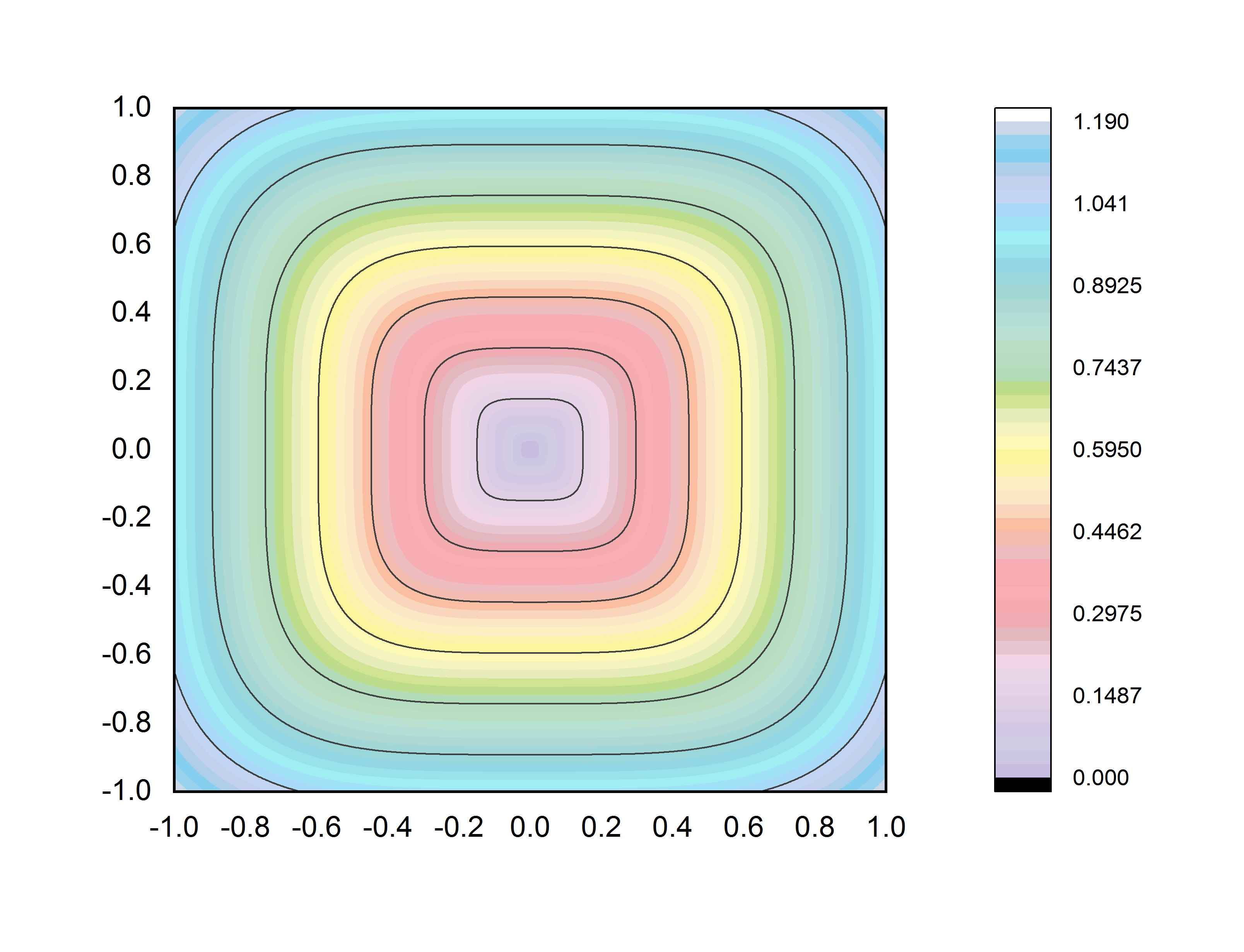}
  \caption{p=4}
  \label{fig:p4}
\end{subfigure}\hfil 
\begin{subfigure}{0.30\textwidth}
  \includegraphics[width=\linewidth]{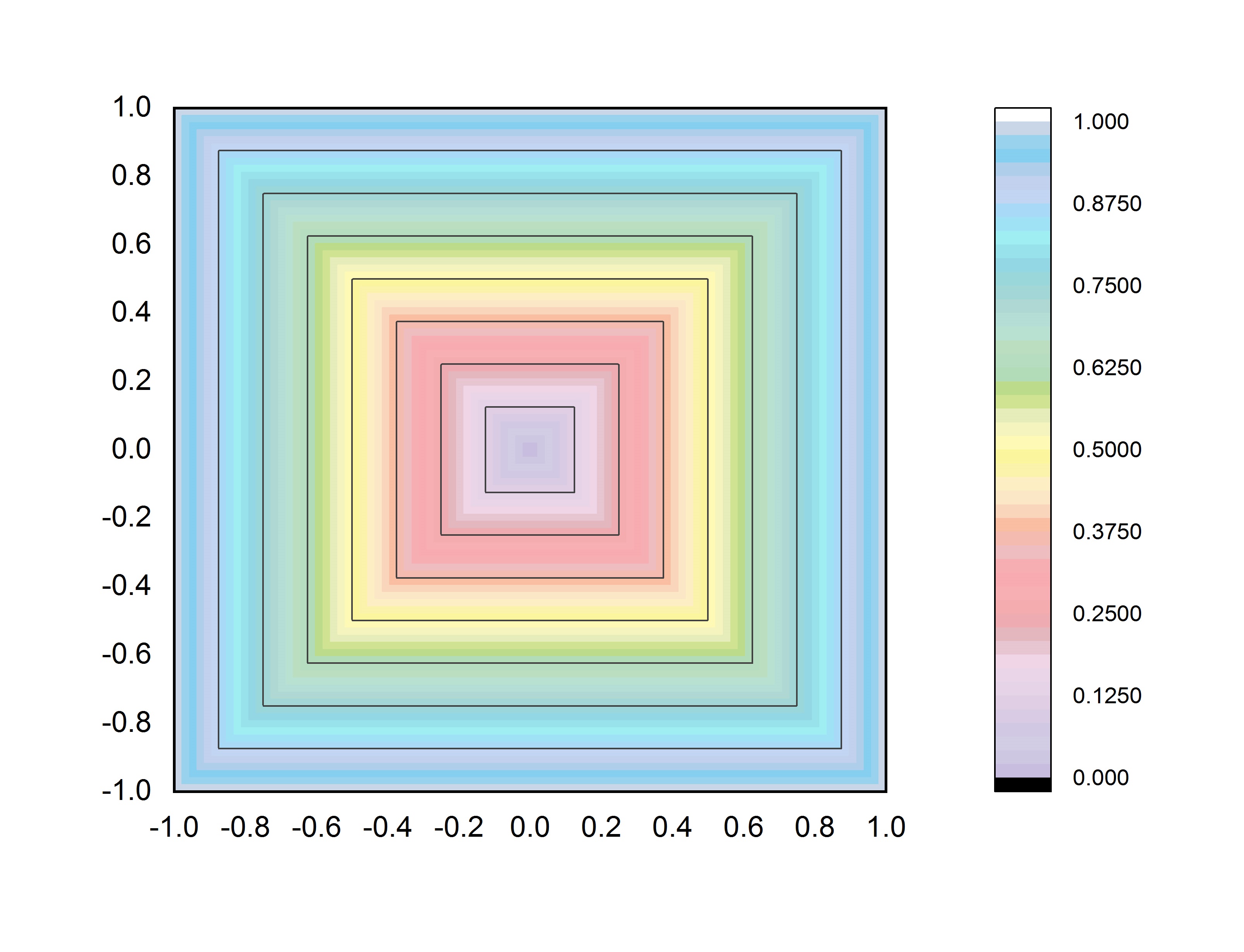}
  \caption{p=$\infty$}
  \label{fig:pInf}
\end{subfigure}
\caption{Isosimilarity contour lines with eight different values of  $p$ in the Minkowski distance formula}
\label{fig:different_P}
\end{figure}

In supervised learning tasks like classification, the value of $p$ could be chosen to maximize model accuracy. However, we do not have that access to the class labels in unsupervised learning. In that case, a sensible way of choosing $p$ could be investigating on use of  $R{V_{F,p}}$ as an objective to be maximized. Also in \cite{flexer2015choosing} the relation between the phenomenon of concentration and hubness has been studied. The authors proposed an unsupervised approach for choosing the value of $p$  by measuring hub and anti-hub occurrence as defined in the paper.
Although  $R{V_{F,p}}$ and hubness can give a sense of the concentration level, they are not excellent measures all the time. As shown in Figure (\ref{fig:different_P}), a different value of $p$ reflects  a distinctive measure of distance in a Euclidean space. It can be seen, the rotation of the coordinate system will also lead to changes in the distance measurement. The only exception is the circular shape, $p=2$ (Figure \ref{fig:p2}). By adopting $p \to 0$ being exact in one dimension has more value than have balance in two (Figure \ref{fig:p_25}). 
conversely, by approaching $p \to \infty $, just the maximum difference is matter (Figure \ref{fig:pInf}). So by going far from $p=2$, the distance meaning completely changes therefore the value of $p$ should be chosen by considering both being meaningful and reduction of the concentration. However, ground truth annotation is often costly and inaccessible in real-world applications, there are usually limited available class labels. In this case, semi-supervised learning provides a better choice by leveraging unlabeled data by using a small set of labeled data. We discover the optimal value of exponent $p$, by uncovering only a few percents of data. We employ the entire data set, either labeled or not to run a series of clustering experiments at various values of $p$ as reported superior results rather than using only labeled data in \cite{de2012minkowski}.
 
\subsubsection{Finding Centers} \label{finding_centers}
After deriving the approximate feature map and an optimal value for $p$ then the kernel clustering algorithm can be started in order to optimize the objective function with considering Minkowski distance. It minimizes the sum of Minkowski distances between instances and the related centers. The Minkowski k-means objective function can be written as below by applying Minkowski distance on (\ref{obj_func}).
\begin{equation}\label{obj_fun_Minkw}
    \arg \mathop {{\rm{min}}}\limits_{M',\delta } \sum\limits_{i = 1}^N {\sum\limits_{k = 1}^K {{\delta _{ik}}||{\Phi _i} - {{m}'_k}|{|^p}} } 
\end{equation}
It should be considered that, it is not possible to use the definition of the center which previously used as Eq (\ref{Euc_center}) since it does not minimize (\ref{obj_fun_Minkw}) when $\delta $  is given constant. In the other words, For finding Minkowski’s centers, we need to find value of vector $m'_k$ for each cluster, minimizes the following function:
\begin{equation}\label{center_Mink_dist}
   \arg \mathop {{\rm{min}}}\limits_{{{m}'_k}} {\rm{ }}{\sum\limits_{{\Phi _i} \in {C_k}} {|{\Phi _i} - {{m}'_k}|} ^p} 
\end{equation}
Being minimizer of (\ref{center_Mink_dist}) is required for vector $m'_k$ in order to prove the convergence of Minkowski K-means to a local optimum. In other words, it lowers the cost in each iteration monotonically. By this way, it is proven that the algorithm will be converged in a finite number of iterations since the algorithm iterates a function whose domain is a finite set and the cost is decreased monotonically.\\\
The search space in order to find  vector $m'_k$ is too large for an exhaustive search because the algorithm needs accurate solutions. Note that finding vector $m'_k$ is a single objective problem and elements on different dimensions are independent. Various evolutionary algorithms can be designed to find the best solutions.  However, for $p>1$ the Eq (\ref{center_Mink_dist}) is a convex function and more desirable algorithms like steepest descent can be used to find the global minimizer \cite{de2012minkowski}.
\subsection{Weighted Version of Explicit Kernel K-means}
Using the feature weighted version of Minkowski distance enables the algorithm to provide a better fit to the cluster structures than is possible with Minkowski K-means alone. It allows our algorithm to find both spherical and non-spherical (elliptical) structural clusters in feature space; accordingly, gives a much better fit to arbitrary shaped clusters in original spaces. Authors in \cite{de2012minkowski}  have extended weighted K-means variants work \cite{chan2004optimization,huang2005automated,huang2008weighting} through transforming feature weights to be feature scale. This means  Minkowski exponent is assigned to feature weights too. The objective function for the weighted version of Minkowski K-means in Equation (\ref{obj_fun_Minkw}) can be written as the following equation:

\begin{equation}
\arg \mathop {{\rm{min}}}\limits_{M',\delta ,W} \sum\limits_{i = 1}^N {\sum\limits_{k = 1}^K {\sum\limits_{l = 1}^{D'} {{\delta _{ik}}{{({w_{kl}})}^p}||\Phi _i^l - m_k^{'l}|{|^p}} } }   \end{equation}
The weight $w_{kl}$ reflects the relevance of feature $l$  at the cluster $k$. They are assigned base on inverse proportion of dispersion of a feature within a certain cluster.  Thus, a feature with small dispersion within a specific cluster would have a higher weight, and vice versa. More precisely, $w_{kl}$ is updated on each iteration as Equation (\ref{weight}), where ${V_{kl}} = \sum\nolimits_{i \in {c_k}} {|\varphi ({x_{iv}}) - {m_{kv}}{|^p}}  $ 
\begin{equation}\label{weight}
    {w_{kv}} = \frac{1}{{\sum\nolimits_{u \in V} {{{(\frac{{{V_{kl}}}}{{{V_{ku}}}})}^{\frac{1}{{p - 1}}}}} }}
\end{equation}
To see the entire algorithm process easily we have also the following flowchart (Fig. \ref{fig:flowchart}): 

\begin{figure}
\caption{The process flowchart of the Explicit kernel MWK-means algorithm}
\label{fig:flowchart}
\centering
\includegraphics[width=0.9\textwidth]{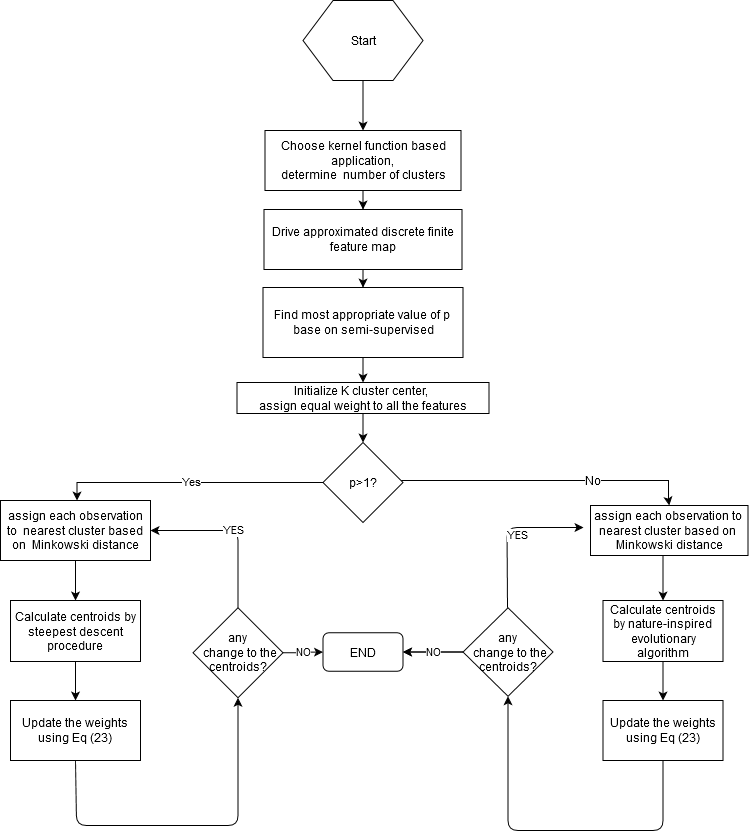}
\end{figure}
\subsection{Accelerate Minkowski  K-means clustering}
The Minkowski's center computation would significantly decelerate the running time. We know already, that at $p=2$ the center is the mean, and at $p=1$ it is given by the component-wise median. Otherwise, it requires an iterative, steepest descent process or an evolutionary computation that can be considered in computation costs. A  significant computational speed-up can be achieved by appropriate initialization in order to lessen the number of iterations and consequently decrease the time of searching for  Minkowski centers as a process of minimization. We use an output vector containing cluster centers of  K-means with $p$ equal 2 or 1  as an initialization. This initialization approach results in an impressive reduction in the time required for clustering, because all of these Minkowski's distances share the same properties and  data  would be half-clustered. In \cite{franti2019much}, the effect of the initialization of K-Means has been studied. The authors found when the clusters overlap, considering various strategies for initialization  would not matter much on the results.
\section{Experimental Results}
In this section, We present the results of performance experiments of our Explicit Kernel MWK-means. We have conducted our experiments on four benchmark data sets: USPS dataset, MNIST dataset, caltech101, and MSRC-V1. Some relevant statistics of them are shown in Table  \ref{tab:data_features}. 
\begin{table}[]
\centering
\caption{Description of the data sets}
\begin{tabular}{|c|c|c|c|}
\hline
             & \# instances & \# features & \# classes \\ \hline
MNIST Digits & 70,000     & 784       & 10       \\ \hline
USPS Digits  & 11,000     & 256       & 10       \\ \hline
Caltech7     & 441        & ~60,000   & 7        \\ \hline
MSRC-V1      & 210        & 68160     & 7        \\ \hline
\end{tabular}
\label{tab:data_features}
\end{table}

Two standard metrics were used to measure the performance of the image clustering that is, Normalized Mutual Information (NMI) and Purity.
\subsection{Data Descriptions}
\textbf{MNIST Dataset} introduced by Yann Lecun and Corinna Cortes. The dataset includes a total of 70,000 samples of digits 0-9. Each sample consists of 28 by 28 pixels which are within the range [ 0,255 ]. \par
\textbf{ USPS dataset} includes 11,000 0-9 digits instances. The dataset is known to be very complicated with a recorded human error rate of 2.5\% . The images are 16 by 16 gray-scale pixels.\par
\textbf{Caltech101 dataset} comprises 9144 images of items belonging to 101 classes and one background class. The number of images in each class differs. The size of each image is approximately 300$\times$200 pixels. We have selected the commonly used 7 groups, i.e. Face, Motorbike, Dolla-Bill, Garfield, Snoopy, Stop-sign and Windsor-chair for our experiments. \par
\textbf{MSRC-V1 dataset} is from Microsoft Research in Cambridge. This data set is commonly used for scene recognition. We adopt Lee and Grauman’s approach \cite{lee2009foreground} to refine and getting 7 classes  including tree, building, airplane, cow, face, car and bicycle where each class has 30 images.
\subsection{Setting of the Experiments}
To figure out which value of the exponent $p$ has the best performance, a semi-supervised manner was employed. The value of $p$ was derived from uncovering  the class of labels on a 15\% data sample though clustering was  conducted over the whole dataset. After running a series of clustering experiments at different values of $p$, the $p$ that produced the higher clustering accuracy was picked. Each of the experiments repeated 50 times and the average NMI is reported.\par 
Images clustering by using raw pixels as features is unlikely to work effectively. The standard practice is to use visual descriptors such as HOG and SIFT. Both of HOG and SIFT use histograms of pixel intensity gradients in their descriptors. The  class of homogeneous kernels are   popular due to its effectiveness. Hog can describe the shape and edge information of  object; and SIFT features are invariant to image scale, rotation, noise and illumination changes. For MNIST and USPS databases the HOG feature is used. We choose $4\times4$ grid cells, $2 \times 2$ cells in one block and 1 cell spacing as the parameter in HOG calculation. For the two other datasets, Caltech7 and MSRC-V1, the key-points are extracted  from each image then represented each keypoint as a 128-dimensional SIFT descriptor. A randomly chosen subset of SIFT features were clustered in order to form a visual vocabulary for either of dataset. Each SIFT descriptor was then quantified into a visual word by considering the nearest cluster center. A 500-dimensional vector representation for each image is obtained.

\subsection{Comparison Results}
Table \ref{tab:semi-learning} clearly shows the match between the value of $p$ learned in semi-supervised and fully supervised settings. The only exception is on MNIST with Using ${\chi ^2}$ kernel that there is no match between learned and optimal value of exponent $p$. However, table \ref{tab:MNIST_experiment} shows it has still better performance than common kernel K-means with the exponent $p=2$.

\begin{table}[h]
\centering
\caption{The effect of $p$ on clustering accuracy results with the semi-supervised learning by  revealing  the class of labels on a 15\%  of data}

\arrayrulecolor{black}
\begin{tabular}{|l|ll|ll|} 
\hline
~                       & \multicolumn{2}{l|}{Exponent p} & \multicolumn{2}{l|}{NMI}  \\ 
\arrayrulecolor{black}\cline{2-5}
~                       & learned & Optimal                & learned & Optimal          \\ 
\arrayrulecolor{black}\hline
MNIST (intersection)    & 1.83   & 1.87                   & 0.7862 & 0.7981           \\ 
\cline{1-1}
MNIST (JS)              & 1.7    & 1.7                    & 0.7692 & 0.7692           \\ 
\cline{1-1}
MNIST(${\chi ^2}$)               & 1.7    & 3.1                    & 0.7772 & 0.7862           \\ 
\hline
USPS (intersection)     & 1.66   & 1.5                    & 0.7212 & 0.7420           \\ 
\cline{1-1}
USPS (JS)               & 1.4    & 1.4                    & 0.7353 & 0.7353           \\ 
\cline{1-1}
USPS (${\chi ^2}$)               & 1.4    & 1.2                    & 0.7291 & 0.7510           \\ 
\hline
Caltech7 (intersection) & 1.1    & 1                      & 0.6558 & 0.6661           \\ 
\cline{1-1}
Caltech7 (JS)           & 1.1    & 0.9                    & 0.6203 & 0.6367           \\ 
\cline{1-1}
Caltech7 (${\chi ^2}$)           & 0.9    & 1.2                    & 0.6547 & 0.6706          \\ 
\hline
MSRC-V1 (intersection)  & 1.22   & 1                      & 0.7228 & 0.7438           \\ 
\cline{1-1}
MSRC-V1 (JS)            & 1.2    & 1.1                    & 0.7358 & 0.7559           \\ 
\cline{1-1}
MSRC-V1 (${\chi ^2}$)\_          & 1.3    & 1.1                    & 0.7565 & 0.7646           \\
\hline
\end{tabular}
\label{tab:semi-learning}
\end{table}

We compared the clustering performance of our method (Explicit kernel MWK-means)  with their respective equal weighted version and constant Euclidean one.  In addition, we also compared the results of our method with the exact kernel K-means clustering. Table \cref{tab:MNIST_experiment,tab:USPS_experiment,tab:Caltech_experiment,tab:MSRC_V1_experiment} demonstrate the clustering results in terms of NMI and purity on all data sets. It can be seen that, compared with common exact kernel K-means counterparts, our proposed Explicit kernel MWK-means  improve the clustering performance on all the data sets. 

\begin{table}[H]
\centering
\caption{MNIST Digits: Adaptive Explicit kernel MWK-means after using HOG descriptor}
\begin{tabular}{|cc|cc|cc|cc|}
\hline
                                                &     & \multicolumn{2}{c|}{intersection} & \multicolumn{2}{c|}{${\chi ^2}$}     & \multicolumn{2}{c|}{JS}     \\
                                                & dm. & NMI              & Purity         & NMI          & Purity       & NMI          & Purity       \\ \hline
\multicolumn{1}{|c|}{Exact Kernel}              & --  & 0.7101$\pm$0.024    & 0.7388$\pm$0.040   & 0.7109$\pm$0.029 & 0.7047$\pm$0.044 & 0.7121$\pm$0.031  & 0.7007$\pm$0.041 \\ \hline
\multicolumn{1}{|c|}{\begin{tabular}[c]{@{}c@{}}Explicit Kernel\\ K-means\end{tabular}}& 3   & 0.7138$\pm$0.023    & 0.7388$\pm$0.040   & 0.7186$\pm$0.028 & 0.7107$\pm$0.045 & 0.7148$\pm$0.033 & 0.7027$\pm$0.042 \\ \hline
\multicolumn{1}{|c|}{\begin{tabular}[c]{@{}c@{}}Explicit kernel\\  MK-means\end{tabular}} & 3  &0.7362$\pm$020       & 0.7380$\pm$0.038   & 0.7192$\pm$0.029 & 0.6868$\pm$0.043 & 0.7272$\pm$0.035 & 0.7127$\pm$0.045 \\ \hline
\multicolumn{1}{|c|}{\begin{tabular}[c]{@{}c@{}}Explicit kernel\\  MWK-means\end{tabular}} & 3   & \textbf{0.7862}$\pm$0.026     & \textbf{0.8048}$\pm$0.039   & \textbf{0.7772}$\pm$0.031 & \textbf{0.7909}$\pm$0.047 & \textbf{0.7692}$\pm$0.038 & \textbf{0.7935}$\pm$0.045 \\ \hline
\end{tabular}
\label{tab:MNIST_experiment}
\end{table}

\begin{table}[H]
\centering
\caption{USPS Digits: Adaptive Explicit kernel MWK-means after using HOG descriptor}
\begin{tabular}{|cc|cc|cc|cc|}
\hline
                                                &     & \multicolumn{2}{c|}{intersection} & \multicolumn{2}{c|}{${\chi ^2}$}     & \multicolumn{2}{c|}{JS}     \\
                                                & dm. & NMI              & Purity         & NMI          & Purity       & NMI          & Purity       \\ \hline
\multicolumn{1}{|c|}{Exact Kernel}              & --  & 0.6567$\pm$0.040    & 0.6981$\pm$0.052   & 0.6695$\pm$0.046 & 0.6932$\pm$0.057 & 0.6506$\pm$0.034  & 0.6976$\pm$0.045 \\ \hline
\multicolumn{1}{|c|}{\begin{tabular}[c]{@{}c@{}}Explicit Kernel\\ K-means\end{tabular}}   & 3   & 0.6591$\pm$0.048     & 0.6986$\pm$0.056   & 0.6631$\pm$0.045 & 0.7006$\pm$0.057 & 0.6621$\pm$0.036 & 0.6923$\pm$0.044 \\ \hline
\multicolumn{1}{|c|}{\begin{tabular}[c]{@{}c@{}}Explicit kernel\\  MK-means\end{tabular}} & 3   & 0.68962$\pm$044      & 0.7087$\pm$0.055   & 0.6953$\pm$0.049 & 0.7107$\pm$0.059 & 0.6991$\pm$0.037 & 0.7029$\pm$0.045 \\ \hline
\multicolumn{1}{|c|}{\begin{tabular}[c]{@{}c@{}}Explicit kernel\\  MWK-means\end{tabular}} & 3   & \textbf{0.7212}$\pm$0.045     & \textbf{0.7578}$\pm$0.055   & \textbf{0.7291}$\pm$0.043 & \textbf{0.7658}$\pm$0.050 & \textbf{0.7353}$\pm$0.039 & \textbf{0.7643}$\pm$0.047 \\ \hline
\end{tabular}
\label{tab:USPS_experiment}
\end{table}

\begin{table}[H]
\centering
\caption{Caltech7: Adaptive Explicit kernel MWK-means  after using SIFT descriptor}
\begin{tabular}{|cc|cc|cc|cc|}
\hline
                                                &     & \multicolumn{2}{c|}{intersection} & \multicolumn{2}{c|}{${\chi ^2}$}     & \multicolumn{2}{c|}{JS}     \\
                                                & dm. & NMI              & Purity         & NMI          & Purity       & NMI          & Purity       \\ \hline
\multicolumn{1}{|c|}{Exact Kernel}              & --  & 0.6064$\pm$0.018    & 0.6581$\pm$0.028   & 0.5918$\pm$0.014 & 0.6849$\pm$0.023 & 0.5649$\pm$0.019 & 0.6710$\pm$0.025 \\ \hline
\multicolumn{1}{|c|}{\begin{tabular}[c]{@{}c@{}}Explicit Kernel\\ K-means\end{tabular}}   & 3   & 0.5961$\pm$0.019     & 0.6941$\pm$0.027   & 0.5981$\pm$0.014 & 0.6858$\pm$0.021 & 0.5919$\pm$0.017 & 0.6735$\pm$0.023 \\ \hline
\multicolumn{1}{|c|}{\begin{tabular}[c]{@{}c@{}}Explicit kernel\\  MK-means\end{tabular}} & 3   & 0.6158$\pm$0.021     & 0.6987$\pm$0.025   & 0.6147$\pm$0.025 & 0.6926$\pm$0.028 & 0.5803$\pm$0.018 & 0.6635$\pm$0.022 \\ \hline
\multicolumn{1}{|c|}{\begin{tabular}[c]{@{}c@{}}Explicit kernel\\  MWK-means\end{tabular}} & 3   & \textbf{0.6558}$\pm$0.024     & \textbf{0.7581}$\pm$0.028   & \textbf{0.6547}$\pm$0.026 & \textbf{0.7419}$\pm$0.029 & \textbf{0.6203}$\pm$0.021 & \textbf{0.7159}$\pm$0.025 \\ \hline
\end{tabular}
\label{tab:Caltech_experiment}
\end{table}

\begin{table}[h]
\centering
\caption{MSRC-v1: Adaptive Explicit kernel MWK-means after using SIFT descriptor}
\begin{tabular}{|cc|cc|cc|cc|}
\hline
                                                &     & \multicolumn{2}{c|}{intersection} & \multicolumn{2}{c|}{${\chi ^2}$}     & \multicolumn{2}{c|}{JS}     \\
                                                & dm. & NMI              & Purity         & NMI          & Purity       & NMI          & Purity       \\ \hline
\multicolumn{1}{|c|}{Exact Kernel}              & --  & 0.6509$\pm$0.024    & 0.7201$\pm$0.029   & 0.6889$\pm$0.028 & 0.7301$\pm$0.031 & 0.6859$\pm$0.034 & 0.7407$\pm$0.039 \\ \hline
\multicolumn{1}{|c|}{\begin{tabular}[c]{@{}c@{}}Explicit Kernel\\ K-means\end{tabular}}   & 3   & 0.6504$\pm$0.022     & 0.7221$\pm$0.023   & 0.6927$\pm$0.030 & 0.7343$\pm$0.034 & 0.6893$\pm$0.035 & 0.7436$\pm$0.040 \\ \hline
\multicolumn{1}{|c|}{\begin{tabular}[c]{@{}c@{}}Explicit kernel\\  MK-means\end{tabular}} & 3   & 0.6728$\pm$0.027     & 0.7256$\pm$0.032   & 0.7058$\pm$0.032 & 0.7390$\pm$0.034 & 0.7065$\pm$0.033 & 0.7541$\pm$0.037 \\ \hline
\multicolumn{1}{|c|}{\begin{tabular}[c]{@{}c@{}}Explicit kernel\\  MWK-means\end{tabular}} & 3   & \textbf{0.7228}$\pm$0.026     & \textbf{0.8105}$\pm$0.030   & \textbf{0.7556}$\pm$0.033 & \textbf{0.8269}$\pm$0.036 & \textbf{0.7358}$\pm$0.030 & \textbf{0.8254}$\pm$0.034 \\ \hline
\end{tabular}
\label{tab:MSRC_V1_experiment}
\end{table}

\section{Conclusions}
In this paper, we proposed a kernel K-means method which is based on explicit feature maps with further matching in feature space. Using adaptive Minkowski metric and feature weighting in feature space enable our algorithm to get high clustering quality and show strong robustness to noise feature and concentration phenomena.  Applying approximate finite-dimensional feature maps were only discussed in the Support Vector Machines (SVM) problems before. We suggested using this method as alleviates storing huge kernel matrix in memory, in addition to access the data explicitly in feature space. Our experiments demonstrate that our proposed method consistently achieves superiors clustering performances  in terms of two standard metrics Normalized Mutual Information (NMI), and Purity, evaluated on four standard bench-mark data sets on object and scene recognition. 
\printbibliography

@article{jiang2004cluster,
  title={Cluster analysis for gene expression data: a survey},
  author={Jiang, Daxin and Tang, Chun and Zhang, Aidong},
  journal={IEEE Transactions on Knowledge \& Data Engineering},
  number={11},
  pages={1370--1386},
  year={2004},
  publisher={IEEE}
}

@inproceedings{daruru2009pervasive,
  title={Pervasive parallelism in data mining: dataflow solution to co-clustering large and sparse netflix data},
  author={Daruru, Srivatsava and Marin, Nena M and Walker, Matt and Ghosh, Joydeep},
  booktitle={Proceedings of the 15th ACM SIGKDD international conference on Knowledge discovery and data mining},
  pages={1115--1124},
  year={2009},
  organization={ACM}
}

@inproceedings{macqueen1967some,
  title={Some methods for classification and analysis of multivariate observations},
  author={MacQueen, James and others},
  booktitle={Proceedings of the fifth Berkeley symposium on mathematical statistics and probability},
  volume={1},
  number={14},
  pages={281--297},
  year={1967},
  organization={Oakland, CA, USA}
}

@inproceedings{inokuchi2004lvq,
  title={LVQ clustering and SOM using a kernel function},
  author={Inokuchi, Ryo and Miyamoto, Sadaaki},
  booktitle={2004 IEEE International Conference on Fuzzy Systems (IEEE Cat. No. 04CH37542)},
  volume={3},
  pages={1497--1500},
  year={2004},
  organization={IEEE}
}

@inproceedings{macdonald2000kernel,
  title={The kernel self-organising map},
  author={MacDonald, Donald and Fyfe, Colin},
  booktitle={KES'2000. Fourth International Conference on Knowledge-Based Intelligent Engineering Systems and Allied Technologies. Proceedings (Cat. No. 00TH8516)},
  volume={1},
  pages={317--320},
  year={2000},
  organization={IEEE}
}

@inproceedings{qin2004kernel,
  title={Kernel neural gas algorithms with application to cluster analysis},
  author={Qin, A Kai and Suganthan, Ponnuthurai N},
  booktitle={Proceedings of the 17th International Conference on Pattern Recognition, 2004. ICPR 2004.},
  volume={4},
  pages={617--620},
  year={2004},
  organization={IEEE}
}

@article{camastra2005novel,
  title={A novel kernel method for clustering},
  author={Camastra, Francesco and Verri, Alessandro},
  journal={IEEE transactions on pattern analysis and machine intelligence},
  volume={27},
  number={5},
  pages={801--805},
  year={2005},
  publisher={IEEE}
}

@article{zhang2004novel,
  title={A novel kernelized fuzzy c-means algorithm with application in medical image segmentation},
  author={Zhang, Dao-Qiang and Chen, Song-Can},
  journal={Artificial intelligence in medicine},
  volume={32},
  number={1},
  pages={37--50},
  year={2004},
  publisher={Elsevier}
}

@inproceedings{zhang2003kernel,
  title={Kernel-based fuzzy and possibilistic c-means clustering},
  author={Zhang, Dao Q and Chen, Song C},
  booktitle={Proceedings of the International Conference Artificial Neural Network},
  volume={122},
  pages={122--125},
  year={2003}
}

@inproceedings{zha2002spectral,
  title={Spectral relaxation for k-means clustering},
  author={Zha, Hongyuan and He, Xiaofeng and Ding, Chris and Gu, Ming and Simon, Horst D},
  booktitle={Advances in neural information processing systems},
  pages={1057--1064},
  year={2002}
}

@inproceedings{ding2005equivalence,
  title={On the equivalence of nonnegative matrix factorization and spectral clustering},
  author={Ding, Chris and He, Xiaofeng and Simon, Horst D},
  booktitle={Proceedings of the 2005 SIAM international conference on data mining},
  pages={606--610},
  year={2005},
  organization={SIAM}
}

@book{dhillon2004unified,
  title={A unified view of kernel k-means, spectral clustering and graph cuts},
  author={Dhillon, Inderjit S and Guan, Yuqiang and Kulis, Brian},
  year={2004},
  publisher={Citeseer}
}

@incollection{dubes1980clustering,
  title={Clustering methodologies in exploratory data analysis},
  author={Dubes, Richard and Jain, Anil K.},
  booktitle={Advances in computers},
  volume={19},
  pages={113--228},
  year={1980},
  publisher={Elsevier}
}

@inproceedings{hinneburg2000nearest,
  title={What is the nearest neighbor in high dimensional spaces?},
  author={Hinneburg, Alexander and Aggarwal, Charu C and Keim, Daniel A},
  booktitle={26th Internat. Conference on Very Large Databases},
  pages={506--515},
  year={2000}
}

@inproceedings{aggarwal2001surprising,
  title={On the surprising behavior of distance metrics in high dimensional space},
  author={Aggarwal, Charu C and Hinneburg, Alexander and Keim, Daniel A},
  booktitle={International conference on database theory},
  pages={420--434},
  year={2001},
  organization={Springer}
}

@article{shawe2000support,
  title={Support vector machines},
  author={Shawe-Taylor, John and Cristianini, Nello},
  journal={An Introduction to Support Vector Machines and Other Kernel-based Learning Methods},
  pages={93--112},
  year={2000},
  publisher={Cambridge University Press}
}

@article{vedaldi2012efficient,
  title={Efficient additive kernels via explicit feature maps},
  author={Vedaldi, Andrea and Zisserman, Andrew},
  journal={IEEE transactions on pattern analysis and machine intelligence},
  volume={34},
  number={3},
  pages={480--492},
  year={2012},
  publisher={IEEE}
}

@article{francois2007concentration,
  title={The concentration of fractional distances},
  author={Francois, Damien and Wertz, Vincent and Verleysen, Michel},
  journal={IEEE Transactions on Knowledge and Data Engineering},
  volume={19},
  number={7},
  pages={873--886},
  year={2007},
  publisher={IEEE}
}

@article{de2012minkowski,
  title={Minkowski metric, feature weighting and anomalous cluster initializing in K-Means clustering},
  author={De Amorim, Renato Cordeiro and Mirkin, Boris},
  journal={Pattern Recognition},
  volume={45},
  title={Choosing lp norms in high-dimensional spaces based on hub analysis},
  pages={1061--1075},
  year={2012},
  publisher={Elsevier}
}

@article{flexer2015choosing,
  title={Choosing lp norms in high-dimensional spaces based on hub analysis},
  author={Flexer, Arthur and Schnitzer, Dominik},
  journal={Neurocomputing},
  volume={169},
  pages={281--287},
  year={2015},
  publisher={Elsevier}
}

@article{chan2004optimization,
  title={An optimization algorithm for clustering using weighted dissimilarity measures},
  author={Chan, Elaine Y and Ching, Wai Ki and Ng, Michael K and Huang, Joshua Z},
  journal={Pattern recognition},
  volume={37},
  number={5},
  pages={943--952},
  year={2004},
  publisher={Elsevier}
}

@article{huang2005automated,
  title={Automated variable weighting in k-means type clustering},
  author={Huang, Joshua Zhexue and Ng, Michael K and Rong, Hongqiang and Li, Zichen},
  journal={IEEE transactions on pattern analysis and machine intelligence},
  volume={27},
  number={5},
  pages={657--668},
  year={2005},
  publisher={IEEE}
}

@article{huang2008weighting,
  title={Weighting method for feature selection in k-means},
  author={Huang, Joshua Zhexue and Xu, Jun and Ng, Michael and Ye, Yunming},
  journal={Computational Methods of feature selection},
  pages={193--209},
  year={2008},
  publisher={Chapman \& Hall, CRC}
}

@article{lee2009foreground,
  title={Foreground focus: Unsupervised learning from partially matching images},
  author={Lee, Yong Jae and Grauman, Kristen},
  journal={International Journal of Computer Vision},
  volume={85},
  number={2},
  pages={143--166},
  year={2009},
  publisher={Springer}
}

@article{hein2004hilbertian,
  title={Hilbertian metrics and positive definite kernels on probability measures},
  author={Hein, Matthias and Bousquet, Olivier},
  year={2004},
  publisher={Max Planck Institute for Biological Cybernetics}
}

@article{franti2019much,
  title={How much can k-means be improved by using better initialization and repeats?},
  author={Fr{\"a}nti, Pasi and Sieranoja, Sami},
  journal={Pattern Recognition},
  volume={93},
  pages={95--112},
  year={2019},
  publisher={Elsevier}
}
\end{document}